\documentclass[runningheads]{include/eccv/llncs}
\usepackage[year=2026,ID=11422]{include/eccv/eccv}
\usepackage{include/eccv/eccvabbrv}
\usepackage{graphicx}
\usepackage{booktabs}
\usepackage{caption}
\usepackage{wrapfig}
\usepackage{subcaption}
\usepackage[normalem]{ulem}

\usepackage[table]{xcolor}
\usepackage[accsupp]{axessibility} 
\usepackage[pagebackref,breaklinks,colorlinks,citecolor=eccvblue]{hyperref}
\usepackage{hyperref}
\usepackage{bbm}

\begin{document}

\title{Rapidly Deploying On-Device Eye Tracking by Distilling Visual Foundation Models}

\titlerunning{DistillGaze}
\newcommand\themethod{DistillGaze{}}

\author{%
Cheng Jiang\inst{1,2}\thanks{Work done during an internship at Meta.}\and
Jogendra Kundu\inst{1}\and
David Colmenares\inst{1}\and
Fengting Yang\inst{1}\and\\
Joseph P Robinson\inst{1}\and
Ali Behrooz\inst{1}\and
Yatong An\inst{1}\thanks{Correspondence: \email{yatong@meta.com}.}
}

\authorrunning{C. Jiang et al.}

\institute{
Meta Reality Labs, USA
\and
University of Michigan, USA
}

\maketitle

\begin{abstract}
Eye tracking (ET) plays a critical role in augmented and virtual reality applications. However, rapidly deploying \emph{high-accuracy}, on-device gaze estimation for new products remains challenging because hardware configurations (e.g., camera placement, camera pose, and illumination) often change across device generations. Visual foundation models (VFMs) excel on natural-image benchmarks and offer a promising path to rapid training and deployment; yet, we find that off-the-shelf VFMs still struggle to reach high accuracy on specialized near-eye infrared images. 
To close this gap, we introduce \emph{\themethod{}}, a framework that distills a VFM using labeled synthetic data and unlabeled real data for rapid, high-accuracy on-device gaze estimation. \themethod{} proceeds in two stages. First, we adapt a VFM into a domain-specialized teacher using synthetic gaze labels and unlabeled real images. Synthetic data provide scalable, high-quality gaze supervision, while unlabeled real data bridges the synthetic-to-real domain gap. Second, we train an on-device student from both teacher guidance and self-training. 
Evaluated on a large-scale crowd-sourced dataset spanning more than 2{,}000 participants, \themethod{} reduces median gaze error by 58.6\% relative to synthetic-only baselines while maintaining a lightweight 256K-parameter model suitable for real-time on-device deployment. More broadly, \themethod{} offers an efficient path to training and deploying ET models that adapt to hardware changes, and a recipe for combining synthetic supervision with unlabeled real data in on-device regression tasks.
\end{abstract}

\section{Introduction}

\begin{figure}[h!t]
\centering
\begin{subfigure}[b]{0.37\textwidth}
    \centering
    \includegraphics[width=\textwidth]{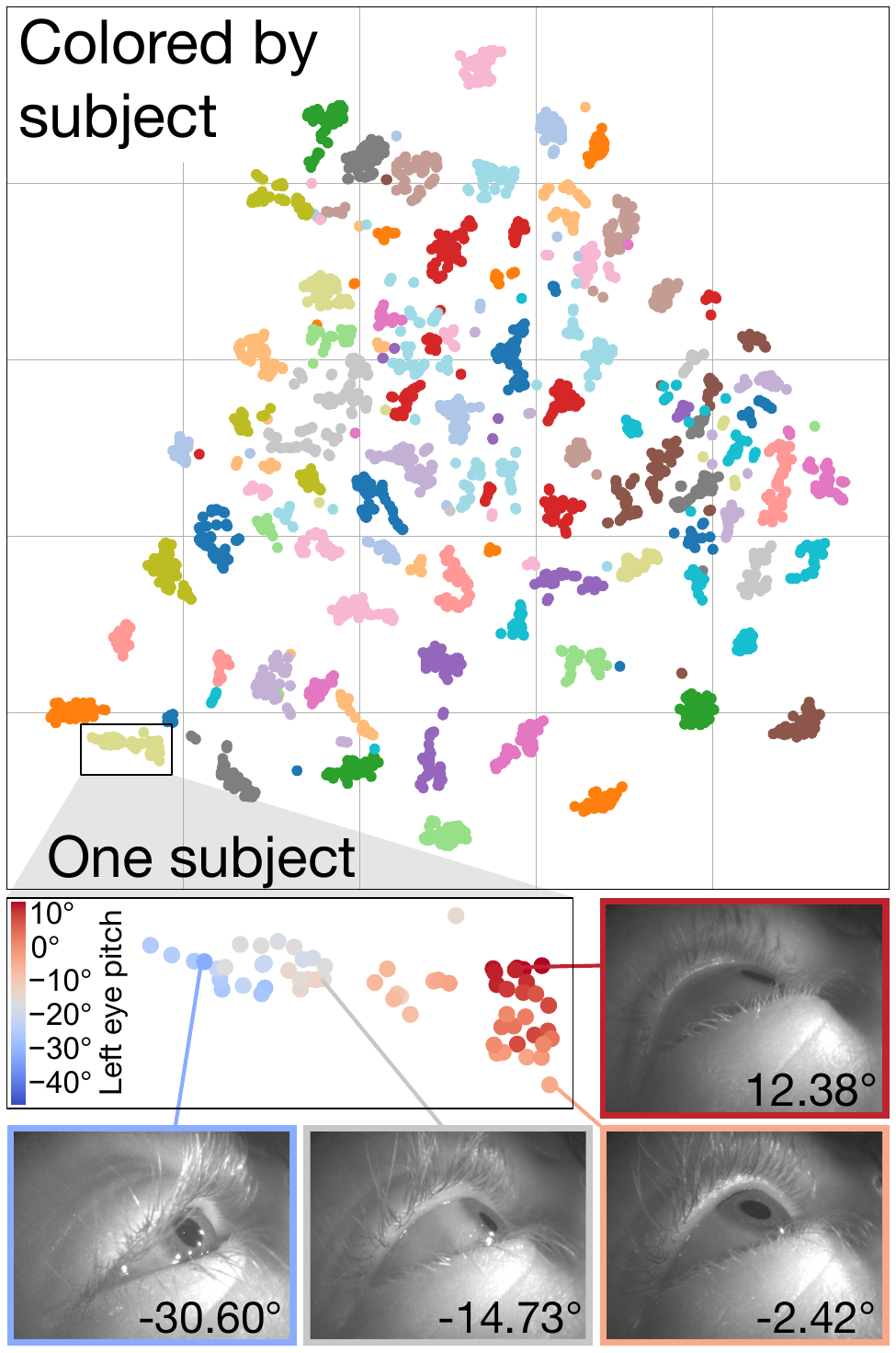}
    \caption{DINOv3 embedding \emph{t}-SNE.}
    \label{fig:motivation.tsne}
\end{subfigure}
\hfill
\begin{subfigure}[b]{0.61\textwidth}
    \includegraphics[width=\textwidth]{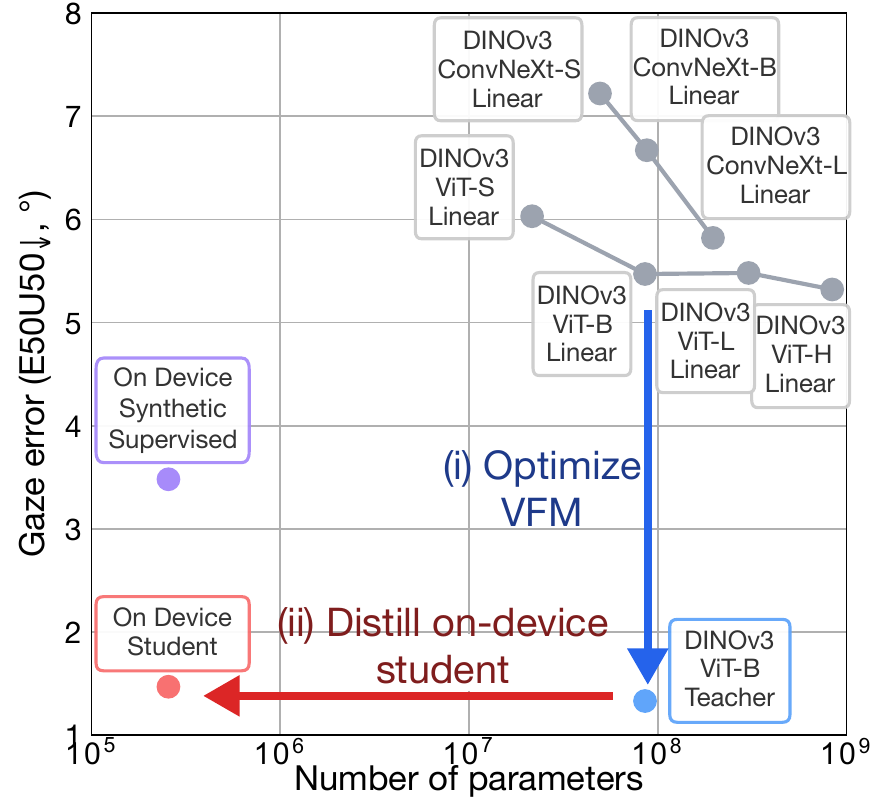}
    \caption{Workflow overview.}
    \label{fig:motivation.overview}
\end{subfigure}
\caption{\textbf{(a) VFM embeddings cluster by subject, obscuring gaze structure.} \emph{t}-SNE of DINOv3 embeddings for near-eye images from 64 randomly sampled subjects: embeddings cluster by subject rather than gaze, yet smooth within-cluster gradients correlate with gaze direction, indicating that gaze cues are present but dominated by identity. \textbf{(b) Workflow overview.} Linear probing of DINOv3 with synthetic supervision (gray) substantially underperforms our on-device baseline (purple), despite having orders of magnitude more parameters, motivating a two-stage approach: (i) optimize a ViT-B teacher from DINOv3; (ii) distill to an on-device student.}
\label{fig:motivation}
\end{figure}

Eye tracking (ET) is a core sensing capability for augmented and virtual reality (AR/VR) applications, enabling foveated rendering, gaze-driven interaction, and attention-aware interfaces~\cite{lin2025digitally}. However, building a production-grade ET system is costly and time-consuming. Conventionally, teams must iterate through hardware prototyping, calibration, data collection, and annotation to obtain ground-truth gaze supervision~\cite{mansour2025enabling}. Each new device generation can trigger a full repetition of data acquisition and model retraining~\cite{lin2025digitally}, due to changes in hardware configuration such as differences in sensor characteristics, camera placement, optics, illumination or on-device image processing. Moreover, gaze supervision itself is often unreliable: even controlled protocols suffer from imperfect fixation and incomplete compliance, producing label noise that is difficult to detect at scale~\cite{mansour2025enabling}. 

As AR/VR hardware iterations accelerate, ET models must be deployed on each new device without waiting for a full labeled-data collection cycle, which for crowd-sourced protocols takes \emph{weeks to months} per device variant. Synthetic data offers a practical path: from a device's camera intrinsics/extrinsics and illumination geometry, photo-realistic near-eye images with pixel-perfect gaze labels can be rendered at scale across diverse eye appearances and gaze directions. Because existing 3D eye assets and subject appearance models are reusable across generations, only the known optical parameters need updating, cutting turnaround from \emph{months to days}. Rendered images, however, do not perfectly match the real target distribution, leaving residual gaps in texture, noise characteristics, and optical artifacts. Unlabeled real images help close this gap at minimal cost: they capture target-device appearance statistics and can be collected passively during normal use, before a calibrated ET system exists. We therefore pair labeled synthetic data with unlabeled real images for \emph{rapid, label-free adaptation} to each new device.

Visual foundation models (VFMs) such as DINOv3~\cite{simeoni2025dinov3} and SAM3~\cite{carion2025sam} have substantially improved representation quality and transfer on natural-image benchmarks, suggesting a path to bypass costly labeled-data collection for eye tracking. These gains, however, are less consistent for specialized imaging modalities, where both the sensing modality and downstream objective differ from web images. This has prompted domain-specific foundation models for medical imaging~\cite{chen2024towards, kondepudi2025foundation}, remote sensing~\cite{liu2024remoteclip, hong2024spectralgpt}, and autonomous driving~\cite{wang2025omnidrive}. On-device ET poses the same challenge: images are captured in near-infrared at close range and contain sensor-, device-, and subject-specific nuances, and high accuracy demands precise regression of subtle appearance variations in the pupil, iris, and eyelids rather than high-level semantic understanding. \cref{fig:motivation} illustrates this limitation. Linear probing of off-the-shelf DINOv3 features (\cref{fig:motivation.overview}, gray) yields gaze errors exceeding 5$^\circ$, far short of task-specific on-device baselines trained on labeled synthetic data (purple) and of on-device ET requirements. The \emph{t}-SNE visualization in \cref{fig:motivation.tsne} explains the gap: representations cluster tightly by subject, with gaze subordinated to identity. Yet, the smooth within-cluster gradients that correlate with gaze direction show that VFMs do encode a usable gaze signal; extracting it requires in-domain adaptation to reshape representations toward gaze-relevant features.

These observations motivate \emph{\themethod{}}, a framework that optimizes a VFM for rapid, label-free ET deployment. Our approach proceeds in two stages (\cref{fig:motivation.overview}). In the first stage (blue arrow), we adapt an off-the-shelf VFM to the eye tracking domain using labeled synthetic images and unlabeled real data, without any real ground-truth gaze labels, to bridge the gap between web-scale pretraining and task-specific ET requirements. In the second stage (red arrow), we distill the optimized model into a lightweight student with two complementary objectives: representation alignment with the fine-tuned VFM teacher and consistency regularization via an exponential moving average (EMA) of the student. The distilled model preserves high-quality ET performance while remaining suitable for real-time on-device deployment.

We evaluate \themethod{} on a large-scale ET dataset collected with Project Aria glasses~\cite{mansour2025enabling} from over 2,000 crowd-sourced participants under in-the-wild conditions, using no ground-truth gaze labels on real data. Our optimized VFM improves both median and tail error over synthetic-only baselines: E50U50 (\ie, the median gaze error of the median subject) drops from 3.48$^\circ$ to 1.33$^\circ$ ($\downarrow$61.8\%), and E90U90 (\ie, the 90th-percentile error of the 90th-percentile users) drops from 14.84$^\circ$ to 8.32$^\circ$ ($\downarrow$43.9\%). The final distilled model achieves 1.44$^\circ$ E50U50 and 8.45$^\circ$ E90U90, 58.6\% and 43.1\% better than the synthetic-supervised baseline, with just 256K parameters, suitable for real-time on-device deployment.

The main contributions of this paper are as follows:
\begin{itemize}
\item We analyze off-the-shelf DINOv3 representations via \emph{t}-SNE and find that embeddings cluster by subject, which limits direct transfer performance on ET. However, intra-subject distributions exhibit smooth gaze-correlated structure, indicating strong potential for adaptation.
\item We propose \emph{\themethod{}}, a two-stage framework that adapts a VFM using synthetic labels and unlabeled real data, without any real gaze labels, then distills it into a compact student for on-device deployment.
\item We achieve state-of-the-art results on a large-scale Project Aria eye tracking benchmark, with significant gains over synthetic-supervised baselines with no increase in model size.
\end{itemize}

\section{Related Work}
\subsection{ET in AR/VR}

ET is essential for AR/VR applications. Typically, ET takes a near-eye image as input and predicts the direction of gaze (\ie pitch and yaw angles).  Eye tracking methods generally fall into two categories: geometry-based and learning-based. Geometric methods assume an explicit eye model (e.g., sphere-on-sphere)~\cite{guestrin2006general,zhu2005eye,liu20203d} and  optimize gaze parameters to align model predictions with image observations. While interpretable and efficient, these methods often make simplifying assumptions about eye shape, struggling with large gaze angles and occlusion. 

Learning-based methods are now dominant, with CNN and transformer architectures achieving improved robustness and accuracy. However, these methods are typically fully supervised, requiring large amounts of annotated gaze data to generalize across users and conditions. Much prior work has been developed for webcams or screen-mounted cameras that capture near-frontal eye views under relatively controlled illumination ~\cite{zhang2015appearance, zhang2020eth, cheng2022gaze, cheng2022puregaze, kellnhofer2019gaze360}, where large-scale data collection is comparatively straightforward. In contrast, head-mounted infrared (IR) cameras in AR/VR headsets operate at extreme off-axis angles with a limited field of view, producing images with perspective distortion, partial iris visibility, and strong specularities due to active IR illumination~\cite{kim2019nvgaze, palmero2020openeds2020, fuhl2021teyed, xiao20252gaze, mansour2025enabling}, making both data collection and generalization substantially more challenging. Compounding this difficulty, even controlled calibration protocols suffer from imperfect fixation and incomplete user compliance, introducing supervision noise that is difficult to detect at scale~\cite{mansour2025enabling}.

To address data acquisition and label quality challenges, synthetic data and domain adaptation methods have gained interest in recent years. Synthetic eye images~\cite{porta2019u2eyes,wood2016learning} can be generated at scale, covering subject diversity and imaging conditions that are impractical to capture with calibrated ground truth~\cite{lin2025digitally}. Yet, the synthetic-to-real domain gap limits transfer performance, which caps accuracy when training relies solely on rendered data. Domain adaptation offers a complementary strategy to bridge the synthetic-to-real gap~\cite{shrivastava2017learning, wood2016learning, nguyen2024deep, wood2015rendering, li2022eyenerf, kim2019nvgaze}. More recently, self-supervised learning has emerged as a paradigm for training encoders on unlabeled data, improving cross-domain generalization~\cite{xu2019self, albuquerque2020improving}.

\subsection{Self-Supervised Representation Learning}
Self-supervised learning (SSL) learns representations by encouraging invariance across augmented views of unlabeled data. Modern SSL methods can be categorized into four broad families~\cite{balestriero2023cookbook}: contrastive learning, self-distillation, canonical correlation analysis, and masked image modeling. Early contrastive methods~\cite{chen2020simple, he2020momentum} require large batch sizes or memory banks, limiting scalability. Self-distillation methods typically adopt a student-teacher framework in which the teacher is an exponential moving average (EMA) of the student to prevent representational collapse~\cite{grill2020bootstrap, caron2021emerging, oquab2023dinov2, simeoni2025dinov3}, and have achieved state-of-the-art performance on a range of benchmarks. Canonical-correlation-analysis-based methods, such as~\cite{bardes2021vicreg}, avoid representational collapse through variance-invariance-covariance regularization without imposing a categorical structure on the embedding space. Masked image modeling methods~\cite{he2022masked, bao2021beit, zhou2021ibot, oquab2023dinov2} take inspiration from language modeling and predict masked tokens from images. Most SSL methods are evaluated on classification or segmentation benchmarks, and many use loss functions that encourage a soft clustering of representations. When extended to large web-scale datasets, these techniques yield visual foundation models (VFMs). Although SSL has transformed representation learning for natural images, its application to near-eye images and gaze estimation remains underexplored.

\subsection{Visual Foundation Models and Knowledge Distillation}
Foundation models pretrained on large-scale data exhibit strong transfer to diverse downstream tasks. CLIP \cite{radford2021learning} aligns visual and language representations through contrastive pretraining on web-scale image-text pairs.
SAM3~\cite{carion2025sam} provides segmentation conditioned on text and visual prompts. Sapiens2~\cite{khirodkarsapiens2} introduces a family of human-centric foundation models, trained on 1-billion images, to learn human appearance and semantics by combining masked reconstruction and contrastive objectives. DINOv3~\cite{simeoni2025dinov3} scales self-supervised learning to a family of models of different sizes, producing features with improved semantic structure. Despite success on natural images, we find these models transfer poorly to near-eye gaze estimation due to the significant domain shift between web-scale RGB images and the infrared, off-axis imagery of AR/VR headsets.

Beyond domain shift, standard VFMs require substantial computational resources to achieve their best performance, rendering them impractical for on-device deployment even at the smallest model size within their family. Knowledge distillation (KD) offers a natural remedy by transferring learned representations from a larger teacher to a compact student network. Early KD methods~\cite{hinton2015distilling, romero2015fitnetshintsdeepnets, zagoruyko2016paying, tung2019similarity, tian2020contrastive} laid the foundation for modern approaches by using soft target logits, intermediate feature matching, attention map alignment, relational structure preservation, and contrastive objectives. Community KD~\cite{lee2023co} trains multiple online students jointly through co-distillation with a pretrained teacher, allowing for bidirectional knowledge transfer. We build on Community KD, adapting it to combine supervision from an optimized teacher and self-distillation of the student.

\begin{figure}[t!]
    \centering

    \begin{subfigure}[t]{\textwidth}
        \centering
        \includegraphics[width=\textwidth]{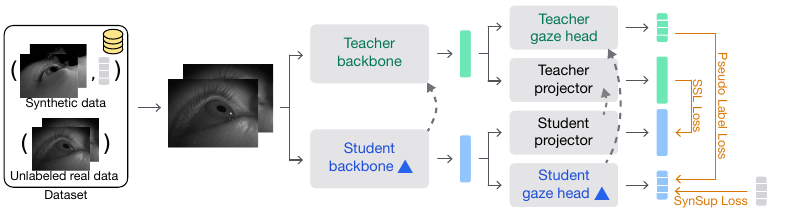}
        \caption{Optimize VFM representations.}
        \label{fig:workflow.optimize}
    \end{subfigure}

    \begin{subfigure}[t]{\textwidth}
        \centering
        \includegraphics[width=\textwidth]{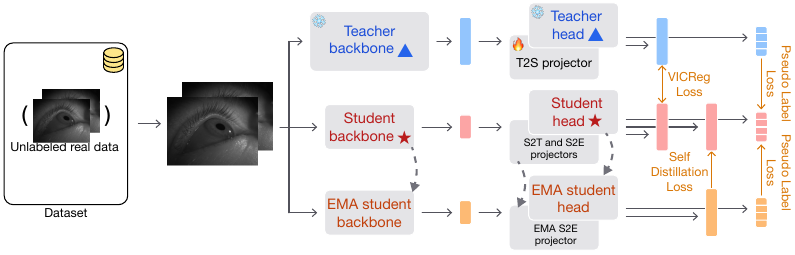}
        \caption{Distill to on-device ET model.}
        \label{fig:workflow.distill}
    \end{subfigure}

    \includegraphics[width=\textwidth]{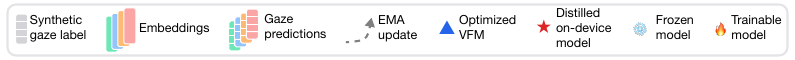}

    \caption{\textbf{Overview of \themethod{}.}
    \textbf{(a)} We optimize a VFM using synthetic supervision and self-distillation on unlabeled real data.
    \textbf{(b)} The optimized VFM is distilled to a lightweight student with 256K parameters, with an EMA student providing complementary supervision. Only the student is deployed.}
    \label{fig:workflow}
\end{figure}

\section{Methods}
\label{sec:methods}

\cref{fig:workflow} illustrates our two-stage approach. We first describe our base ET architecture (\cref{sec:e2e_pipeline}), then detail our VFM optimization strategy (\cref{sec:method.alignment}), and finally present the distillation procedure (\cref{sec:method.distill}).

\subsection{Preliminaries: ET Gaze Estimation Model}
\label{sec:e2e_pipeline}
Our gaze estimation architecture follows \cite{mansour2025enabling}, mapping a synchronized pair of eye images to a per-eye gaze direction. Given input images $\boldsymbol{X}=(\boldsymbol{X}^{L},\boldsymbol{X}^{R})$ for the left and right eyes, a shared-weight backbone $f$ extracts per-eye features:
\begin{equation}
\boldsymbol{h}^{L}=f(\boldsymbol{X}^{L}), \quad \boldsymbol{h}^{R}=f(\boldsymbol{X}^{R}),
\end{equation}
which are concatenated to form a binocular representation $\boldsymbol{h}=[\boldsymbol{h}^{L};\boldsymbol{h}^{R}]$. A fully connected regression head $g$ predicts per-eye gaze rays:
\begin{equation}
\hat{\boldsymbol{y}}=g(\boldsymbol{h})=\left[\hat{\theta}^{L},\hat{\varphi}^{L},\hat{\theta}^{R},\hat{\varphi}^{R}\right]\in\mathbb{R}^{4},
\end{equation}
where $\theta$ and $\varphi$ denote yaw and pitch angles, respectively.

Following \cite{mansour2025enabling}, we supervise gaze predictions using a smooth L1 loss with outlier rejection, which provides robustness to annotation noise:
\begin{equation}
\mathcal{L}_\text{Gaze}(\boldsymbol{y}, \hat{\boldsymbol{y}}) = \begin{cases}
    \frac{1}{2}\frac{(\boldsymbol{y} - \hat{\boldsymbol{y}})^2}{\beta}         & \text{if } |\boldsymbol{y} - \hat{\boldsymbol{y}}| < \beta, \\
    |\boldsymbol{y} - \hat{\boldsymbol{y}}| - \tfrac{1}{2} \beta               & \text{if } \beta \leq |\boldsymbol{y} - \hat{\boldsymbol{y}}| < \gamma, \\
    k (|\boldsymbol{y} - \hat{\boldsymbol{y}}| - \tfrac{1}{2}\beta)            & \text{if } |\boldsymbol{y} - \hat{\boldsymbol{y}}| \geq \gamma,
\end{cases}
\end{equation}
where $k$, $\beta$, and $\gamma$ are hyperparameters, and $\boldsymbol{y}$ is normalized gaze supervision. The loss is quadratic for small errors, linear for moderate errors, and scaled by $k < 1$ for large outliers beyond $\gamma$.

\subsection{Optimizing VFM Representations}\label{sec:method.alignment}
Foundation models pretrained on web-scale data generalize well across many vision tasks. However, frozen foundation-model features underperform a small model trained on simulated data for near-eye gaze estimation (\cref{sec:results.vfm}), suggesting that their high-level representations, optimized for semantic understanding, do not transfer directly to our domain, even if low-level features remain useful.

To adapt these representations for gaze estimation, we adopt a teacher-student self-distillation framework combining synthetic supervision with pseudo-label training on unlabeled real data (\cref{fig:workflow.optimize}). Let $f_t$, $f_s$ denote the initialized DINOv3 teacher and student backbones, and $p_t$, $p_s$ their projectors, respectively. The teacher receives weakly augmented images $\boldsymbol{X}_w$ and the student receives strongly augmented images $\boldsymbol{X}_s$. We compute projected embeddings:
\begin{equation}
\boldsymbol{z}_t = p_t(\boldsymbol{h}_t) = p_t(f_t(\boldsymbol{X}_w)), \qquad 
\boldsymbol{z}_s = p_s(\boldsymbol{h}_s) = p_s(f_s(\boldsymbol{X}_s)).
\end{equation}
The self-distillation loss encourages alignment between teacher and student embeddings:
\begin{equation}
\mathcal{L}_{\text{SD}}^\text{VFM} = \| \boldsymbol{z}_t - \boldsymbol{z}_s \|_2^2
\end{equation}
We then predict gaze using the prediction heads $g_t, g_s$ for teacher and student models, respectively. Let $\hat{\boldsymbol{y}}_t = g_t(\boldsymbol{h}_t)$ and $\hat{\boldsymbol{y}}_s = g_s(\boldsymbol{h}_s)$ denote the teacher and student predictions, respectively. For synthetic data with labels $\boldsymbol{y}$, we apply synthetic supervised (SynSup) gaze loss:
\begin{equation}
\mathcal{L}_{\text{SynSup}} = \mathcal{L}_\text{Gaze}(\boldsymbol{y}, \hat{\boldsymbol{y}}_s).
\end{equation}
For unlabeled real data, we use teacher predictions as pseudo-labels:
\begin{equation}
\mathcal{L}_{\text{Pseudo}} = \mathcal{L}_\text{Gaze}(\hat{\boldsymbol{y}}_t, \hat{\boldsymbol{y}}_s).
\end{equation}
The total objective for optimizing the VFM is:%
\begin{equation}
\begin{aligned}
\mathcal{L}_\text{OptVFM}
&=
\lambda_\text{SynSup}
\sum_{\boldsymbol{X}\in\mathcal{D}_{\text{syn}}}
\mathcal{L}_{\text{SynSup}}\\
&+
(1-\lambda_\text{SynSup})\left[
\sum_{\boldsymbol{X}\in\mathcal{D}_{\text{real}}\cup\mathcal{D}_{\text{syn}}}
\mathcal{L}_{\text{SD}}^\text{VFM}
+
\sum_{\boldsymbol{X}\in\mathcal{D}_{\text{real}}}
\mathcal{L}_{\text{Pseudo}}\right],
\end{aligned}
\end{equation}
where $\mathcal{D}_{\text{syn}}$, $\mathcal{D}_{\text{real}}$ denote the labeled synthetic and unlabeled real datasets, respectively. The hyperparameter $\lambda_\text{SynSup}$ balances the two sources of supervision and is annealed using a cosine schedule: training initially assigns a high weight to synthetic supervision to align the VFM representations with the gaze task, then gradually shifts to self-distillation using all available data, along with pseudo-label supervision using unlabeled real data.

For stable self-distillation, the teacher parameters are updated via EMA~\cite{grill2020bootstrap, caron2021emerging}:
\begin{equation}
    \theta_t \leftarrow \alpha \theta_t + (1 - \alpha) \theta_s,
\end{equation}
where $\theta_t$ and $\theta_s$ denote the teacher and student parameters, respectively, including the backbone, projector, and gaze head.

\subsection{Distilling to an On-Device ET Model}\label{sec:method.distill}
With the optimized VFM from \cref{sec:method.alignment}, we now distill its representations into a lightweight student network suitable for on-device deployment. As illustrated in \cref{fig:workflow.distill}, our framework comprises three models:
\begin{itemize}
    \item A large \textbf{teacher} $f_t$: our fine-tuned VFM from \cref{sec:method.alignment};
    \item A \textbf{student} $f_s$: an on-device model pretrained on synthetic data;
    \item An \textbf{EMA student} $f_e$: an exponential moving average of $f_s$.
\end{itemize}

Here, we reuse the teacher/student notation, but importantly, the roles of these models differ from \cref{sec:method.alignment}: in the distillation stage, the teacher is the optimized VFM and the student is an on-device model. We additionally introduce an EMA student to provide complementary targets for self-distillation and pseudo-label supervision. Our design is inspired by community KD \cite{lee2023co} with key differences: (1) we distill in both feature space (via a VIC-KD loss \cite{guimaraes2024vic}) and prediction space (via pseudo-labels); and (2) the EMA student is updated via momentum rather than gradient descent, drawing inspiration from momentum-based self-supervised learning \cite{grill2020bootstrap, caron2021emerging}. All three models share a common structure: a backbone for image feature extraction, projectors that map features into distillation spaces, and a prediction head to estimate gaze. The teacher uses the optimized VFM backbone $f_t$ and gaze head $g_t$ from \cref{sec:method.alignment}; the student models use backbones and heads pretrained on synthetic data. The projectors are randomly initialized. During training, the teacher and the EMA student receive weakly augmented images $\boldsymbol{X}_w$, while the student receives strongly augmented images $\boldsymbol{X}_s$.

Given an image, the teacher produces embedding $\boldsymbol{h}_t = f_t(\boldsymbol{X}_w)$, projection $\boldsymbol{z}_{t} = p_t(\boldsymbol{h}_t)$ for distillation, and gaze prediction $\hat{\boldsymbol{y}}_t = g_t(\boldsymbol{h}_t)$. The student backbone $f_s$ extracts features $\boldsymbol{h}_s = f_s(\boldsymbol{X}_s)$, which are mapped to projections $\boldsymbol{z}_{s \rightarrow t} = p_{s \rightarrow t}(\boldsymbol{h}_s)$ and $\boldsymbol{z}_{s \rightarrow e} = p_{s \rightarrow e}(\boldsymbol{h}_s)$ for distillation with the teacher and the EMA student, respectively. The head $g_s$ produces gaze prediction $\hat{\boldsymbol{y}}_s = g_s(\boldsymbol{h}_s)$. The student receives gradient updates from all loss terms. 

The EMA student maintains the backbone $f_e$, the projector $p_e$, and the head $g_e$, all updated via the exponential moving average from $f_s$, $p_{s\rightarrow e}$, and $g_s$, respectively. This produces stable embedding $\boldsymbol{h}_e = f_e(\boldsymbol{X}_w)$, projection $\boldsymbol{z}_e = p_e(\boldsymbol{h}_e)$, and prediction $\hat{\boldsymbol{y}}_e = g_e(\boldsymbol{h}_e)$.

We train the framework with complementary objectives in both feature space and gaze prediction space. For teacher-student distillation in feature space, we use an invariance loss between teacher and student projections, along with variance and covariance regularization:
\begin{equation}
\mathcal{L}_{\text{KD}} = \lambda_{\text{inv}} \| \boldsymbol{z}_t - \boldsymbol{z}_{s \rightarrow t} \|_2^2 + \lambda_{\text{var}} v(\boldsymbol{z}_{s \rightarrow t}) + \lambda_{\text{cov}}c(\boldsymbol{z}_{s \rightarrow t}),
\end{equation}
where $v(\cdot)$ and $c(\cdot)$ denote variance and covariance regularization terms~\cite{bardes2021vicreg}. $\lambda_{\text{inv}}$, $\lambda_{\text{var}}$, and $\lambda_{\text{cov}}$ are hyperparameters. The teacher also provides pseudo-labels:
\begin{equation}
\mathcal{L}_{\text{Pseudo-}t} = \mathcal{L}_{\text{Gaze}}(\hat{\boldsymbol{y}}_t, \hat{\boldsymbol{y}}_s).
\end{equation}
The EMA student acts as a momentum teacher for self-distillation. We use an L2 alignment loss and pseudo-label supervision:
\begin{align}
\mathcal{L}_{\text{SD}}^\text{On-device}   = \| \boldsymbol{z}_e - \boldsymbol{z}_{s \rightarrow e} \|_2^2; \qquad
\mathcal{L}_{\text{Pseudo-}e} = \mathcal{L}_{\text{Gaze}}(\hat{\boldsymbol{y}}_e, \hat{\boldsymbol{y}}_s).
\end{align}
The total loss to distill an on-device model  is:
\begin{equation}
\mathcal{L_\text{\themethod{}}} = \sum_{\boldsymbol{X}\in\mathcal{D}_{\text{real}}}\left[\lambda_t (\mathcal{L}_{\text{KD}} + \mathcal{L}_{\text{Pseudo-}t}) + (1-\lambda_t) (\mathcal{L}_{\text{SD}}^\text{On-device} + \mathcal{L}_{\text{Pseudo-}e})\right],
\end{equation}
where $\lambda_t$ is a hyperparameter that weights the two sources of supervision: the optimized VFM teacher and the EMA student. A cosine schedule controls their relative contribution: it emphasizes distillation from the optimized VFM early in training and gradually shifts toward EMA self-distillation. This allows the student to first inherit structured representations from the optimized VFM and then refine them using stable EMA targets. At inference time, we deploy only the on-device student network $f_s$.

\section{Experimental design}
\subsection{Dataset}\label{exp.data}
We evaluate our method on data collected with Project Aria~\cite{mansour2025enabling}. Following the Aria convention, our real dataset comprises 6,299 recordings from 2,222 crowd-sourced participants, partitioned into training and validation splits of 1,825 and 397 participants, respectively. To simulate the unlabeled-data setting, we withhold the ground-truth gaze labels for the training set. Input images are captured by two global-shutter monochrome cameras under diffused infrared illumination. The ET cameras record at 20\,fps for approximately 60 seconds per recording; we temporally subsample by a factor of 10 for training efficiency. Each eye camera captures at $640 \times 480$ resolution, which we downsample to $320 \times 240$ following \cite{mansour2025enabling}. The dataset provides aligned 3D gaze targets. We compute gaze rays using a fixed origin derived from a nominal interpupillary distance of 63.5\,mm. Our synthetic data comprise 165K frames from 998 subjects, rendered in Blender.

\subsection{Implementation Details}
Our on-device ET pipeline contains 256K parameters and uses an FBNet~\cite{wu2019fbnet} backbone shared between images of the left and right eyes. We train with AdamW using a cosine learning-rate schedule with a 10\% warm-up period and a batch size of 256. The learning rate is adjusted between $10^{-3}$ and $10^{-5}$, and each experiment trains for up to 50,000 iterations. For self-distillation, the EMA teacher weights are updated every 100 iterations with the momentum adjusted between 0.95 and 0.99. The feature projection head consists of three fully connected layers with GELU activations. For VFM experiments, we initialize from the ViT-B variant of DINOv3~\cite{simeoni2025dinov3} (86M parameters). All experiments were conducted on 4$\times$ NVIDIA A100 80GB GPUs.

\paragraph{Image transformations.} We employ two augmentation pipelines: strong augmentation for the student, and weak augmentation for the teacher and EMA student. The weak pipeline consists of gamma jitter and random scaling. The strong pipeline additionally applies camera and illumination augmentations, each with probability 0.3. Camera augmentations simulate sensor degradation through modulation transfer function (MTF) filtering, motion blur, and compression artifacts, while illumination augmentations comprise brightness and contrast adjustment, glint inpainting, random shadows, and coarse dropout. 

\subsection{Evaluation Protocol}\label{sec:exp.eval}
We measure gaze estimation accuracy as the angular error (in degrees) between the predicted and ground-truth 3D gaze vectors, averaged over the left and right eyes for each frame. Following~\cite{aziz2024evaluation, mansour2025enabling}, we report results using an Error-User (EU) table to capture both frame-level and subject-level performance. We summarize each user's per-frame error distribution via percentiles E50, E75, and E90, then aggregate across users at U50, U75, and U90. Thus, E50U50 reflects the median error of the median user, whereas E90U90 characterizes tail performance. For efficiency, we uniformly sample 64 frames per recording: 9 frames are reserved for test-time personalization, and metrics are computed on the remaining 55.  We report 95\% confidence intervals via a hierarchical bootstrap over users and frames (1,000 iterations).
\newcommand\graycell[1]{\textcolor{black!50}{#1}}

\begin{table}[t!]
    \centering\caption{\textbf{Gaze estimation performance comparing linear-evaluated foundation models and models optimized with synthetic and unlabeled real data.} 95\% confidence intervals in parentheses. Inf.\ params: number of parameters at inference, SynSup: synthetic supervised, SynFT: synthetic fine-tuning.}\label{tab:results.main}%
    \setlength{\tabcolsep}{1ex}
    \resizebox{\textwidth}{!}{
\begin{tabular}{lc ccc}
\toprule
& Inf. params & E50U50 $\downarrow$ &  E75U75 $\downarrow$ & E90U90 $\downarrow$\\
\midrule
On-device SynSup {\color[HTML]{7c3aed}$\blacklozenge$}  & 256 K & 3.48 (0.19)   & 7.41 (0.53)    & 14.84 (2.03) \\
DINOv3~\cite{simeoni2025dinov3} Linear Probe            & 86 M  & 5.47 (0.23)   & 10.64 (0.61)   & 18.74 (1.60) \\
DINOv3~\cite{simeoni2025dinov3} SynFT                   & 86 M  & 2.01 (0.11)   & 4.29 (0.46)    & 12.11 (1.50) \\
DARE-GRAM~\cite{nejjar2023dare}                         & 86 M  & 4.84 (0.24)   & 9.36 (0.54)    & 17.89 (1.65) \\
Optimized VFM {\color[HTML]{2563eb}$\blacktriangle$}    & 86 M  & \textbf{1.33 (0.07)}   & \textbf{3.07 (0.34)}    & \textbf{8.32 (2.10)}  \\\bottomrule
\end{tabular}
}
\end{table}

\section{Results} \label{sec:results}

We first evaluate DINOv3 VFMs in \cref{sec:results.vfm}, then present our optimized VFM performance in \cref{sec:results.teacher}, the on-device distillation in \cref{sec:results.distill}, and ablation studies for our training dynamics in \cref{sec:results.ablate}.

\subsection{Performance of Off-the-Shelf VFMs on Eye Tracking} \label{sec:results.vfm}

Given the strong transfer performance of VFMs across diverse vision tasks, we first evaluate whether off-the-shelf representations directly benefit synthetic supervised gaze estimation. We extract embeddings from the ViT-B variant of the DINOv3 family~\cite{simeoni2025dinov3}, fit a linear regressor on synthetic data, and apply our per-subject calibration protocol (\cref{sec:exp.eval}). 

As shown in the first section of \cref{tab:results.main}, DINOv3 with linear probe trained on synthetic data yields substantially lower accuracy than our small on-device model trained with synthetic supervision (on-device SynSup), despite having 336$\times$ more parameters.  This observation is consistent across all DINOv3 variants, spanning both ViT and ConvNeXt architectures, as illustrated in  \cref{fig:motivation.overview}. Fine-tuning DINOv3 ViT-B on synthetic data (DINOv3 SynFT) substantially improves performance (\cref{tab:results.main}), reducing E50U50 from 5.47$^\circ$ to 2.01$^\circ$ and surpassing the on-device baseline by leveraging the VFM prior and larger model capacity. This confirms our hypothesis: VFMs provide a useful initialization but require further optimization on domain-relevant data for ET.

\subsection{Optimize VFM Teacher for Eye Tracking}\label{sec:results.teacher}
We next study how to adapt the VFM teacher for eye tracking using synthetic labeled data and unlabeled real data. Unsupervised domain adaptation (UDA) is a natural way to leverage synthetic data for VFM optimization. DARE-GRAM~\cite{nejjar2023dare} is a state-of-the-art UDA method designed for regression tasks. However, as shown in \cref{tab:results.main}, it performed worse than DINOv3 SynFT, potentially due to overfitting to the synthetic data distribution, which may amplify the mismatch between synthetic and real data. 

In contrast, we optimize the VFM with the combined self-distillation and synthetic supervision objective described in \cref{sec:method.alignment}. The result is shown in \cref{tab:results.main}. Compared to DINOv3 SynFT, our optimized VFM reduces E50U50 from 2.01$^\circ$ to 1.33$^\circ$ (33.8\% improvement). The gains are consistent in the tail: E90U90 decreases from 12.11$^\circ$ to 8.32$^\circ$ (31.3\% improvement). In particular, our optimized VFM substantially outperforms the synthetic on-device baseline (1.33$^\circ$ vs.\ 3.48$^\circ$ at E50U50, a 61.8\% reduction). This shows that self-supervised learning on unlabeled real data can effectively bridge the synthetic-to-real gap. Our optimized ViT-B VFM serves as the distillation source in \cref{sec:results.distill}, since our ultimate goal is a deployable on-device model.

\begin{table}[t!]
    \centering\caption{\textbf{On-device model performance comparison.} 95\% confidence intervals in parentheses. Fully supervised upper bound is provided as reference, as it is trained on data that would not be available at the time of product development (bold = best among on-device students). SynSup: synthetic supervised, Inf. params: number of parameters at inference time.}
    \label{tab:results.distill}
    \setlength{\tabcolsep}{1ex}
    \resizebox{\textwidth}{!}{
\begin{tabular}{lc ccc}
\toprule
& Inf. params & E50U50 $\downarrow$ &  E75U75 $\downarrow$ & E90U90 $\downarrow$\\
\midrule
\multicolumn{5}{l}{\emph{Methods without VFM}}\\
On-device SynSup {\color[HTML]{7c3aed}$\blacklozenge$} & 256 K & 3.48 (0.19) & 7.41 (0.53) & 14.84 (2.03)\\
DARE-GRAM~\cite{nejjar2023dare}                                       & 256 K & 2.96 (0.17) & 6.20 (0.47) & 13.41 (1.73)\\
Self-distillation                                                     & 256 K & 2.20 (0.11) & 4.64 (0.46) & 10.95 (2.10)\\\midrule
\multicolumn{5}{l}{\emph{Distillation with Optimized VFM}}\\
Optimized VFM {\color[HTML]{2563eb}$\blacktriangle$} & 86 M  & 1.33 (0.07) & 3.07 (0.34) & 8.32 (2.10)\\
Pseudo labels                                        & 256 K & 1.59 (0.09) & 3.49 (0.30) & \textbf{8.19 (1.56)}\\
SP~\cite{tung2019similarity}                         & 256 K & 1.59 (0.09) & 3.50 (0.30) & 8.21 (1.71)\\
VIC-KD~\cite{guimaraes2024vic}                       & 256 K & 1.46 (0.08) & 3.32 (0.35) & 8.40 (2.01)\\
Community KD~\cite{lee2023co}                        & 256 K & 1.48 (0.09) & \textbf{3.24 (0.31)} & 8.47 (1.88)\\
\themethod{} (ours) {\color[HTML]{dc2626}$\bigstar$} & 256 K & \textbf{1.44 (0.09)} & 3.29 (0.32) & 8.45 (2.04)\\\midrule
\graycell{Fully supervised (upper bound)} & \graycell{256 K} & \graycell{0.82 (0.05)} & \graycell{1.94 (0.26)} & \graycell{5.91 (1.61)}\\
\bottomrule
\end{tabular}
}
\end{table}

\begin{figure}[t!]
    \centering
    \includegraphics[width=\textwidth]{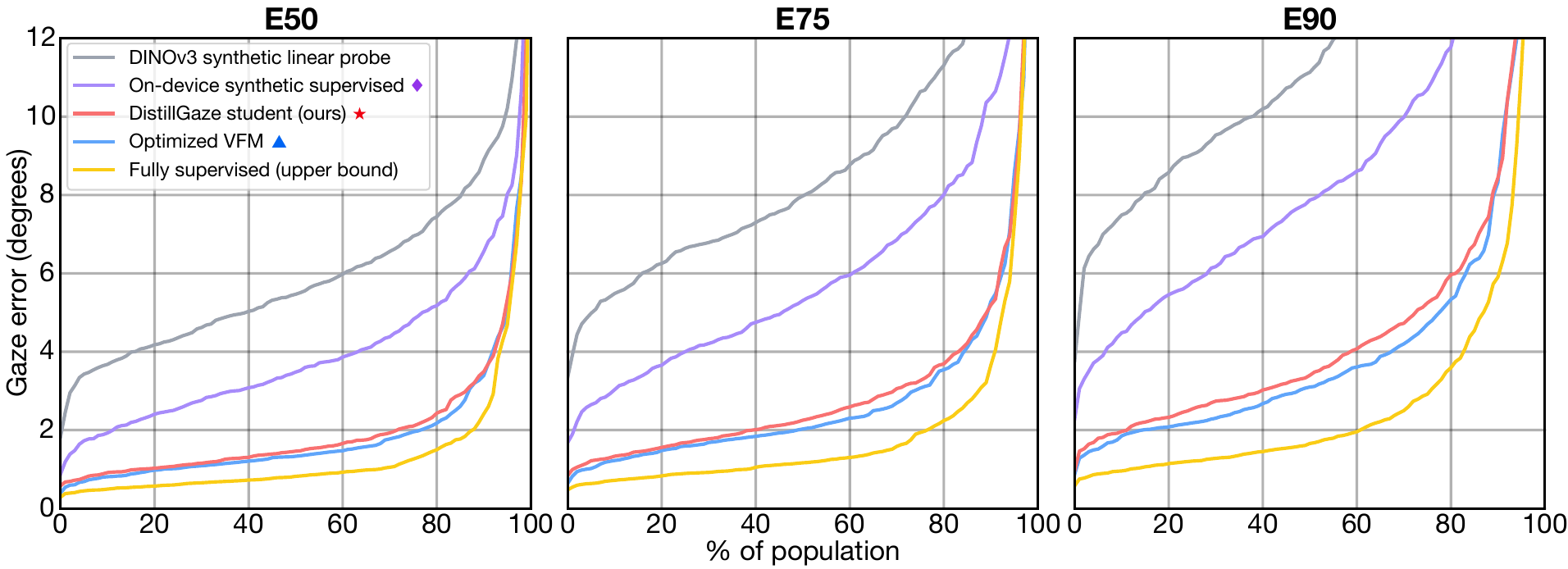}
    
    \caption{\textbf{Population coverage across error percentiles.} We plot the cumulative distribution of users at E50, E75, and E90. Our optimized VFM and \themethod{} student exhibit closely matched performance, demonstrating effective knowledge transfer. Both methods approach the fully supervised on-device upper bound (yellow).
}
    \label{fig:coverage}
\end{figure}

\subsection{Distill to On-Device Model} \label{sec:results.distill}
We evaluate our distillation framework (\cref{sec:method.distill}) to transfer knowledge from our optimized VFM to an on-device ET model. We compare \themethod{} against DARE-GRAM~\cite{nejjar2023dare}, a state-of-the-art UDA method for adaptation without teacher, and several knowledge distillation approaches utilizing the teacher, including VIC-KD~\cite{guimaraes2024vic} and Community KD~\cite{lee2023co}. The results are summarized in \cref{tab:results.distill}.

\paragraph{Methods without VFM.} To demonstrate the value of optimizing the VFM, we trained the small, on-device model without VFM teacher, using synthetic supervision and unlabeled real data. We conducted two experiments. First, we trained DARE-GRAM, a state-of-the-art UDA method. This reduced E50U50 to 2.96$^\circ$, outperforming training with synthetic supervision. Second, we conducted self-distillation following \cref{sec:method.alignment}, using an on-device model, which reduces E50U50 to 2.20$^\circ$, a 36.8\% improvement. This shows that training with unlabeled real data significantly improves model performance even without a VFM teacher, but the remaining gap to the optimized VFM highlights the importance of the VFM prior and learning capacity.

\begin{figure}[t!]
    \centering
    \includegraphics[width=\textwidth]{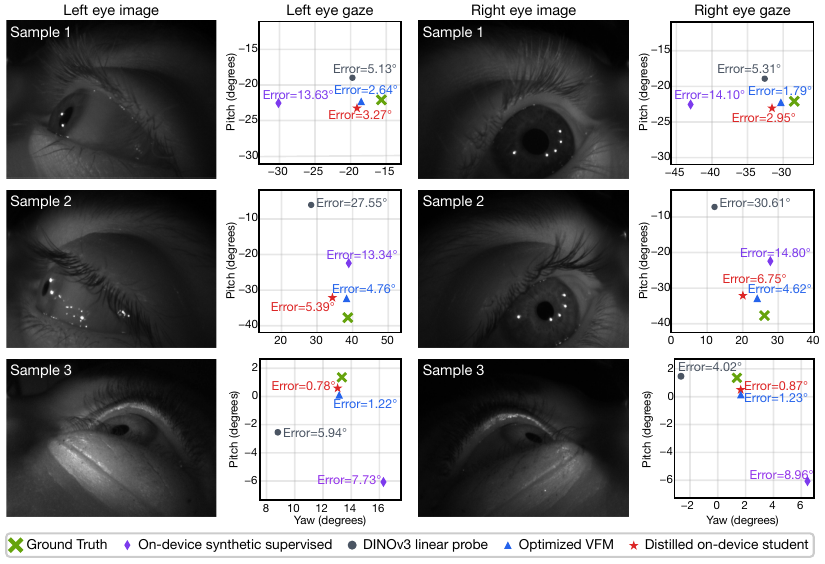}
    \caption{\textbf{Qualitative comparisons of gaze predictions.} Pitch-yaw plots (degrees) compare ground truth (green) against four methods. The on-device synthetic supervised model (purple) and the frozen DINOv3 linear probe (gray) yield large errors, highlighting the need for adaptation beyond synthetic-only training or frozen VFM features. Our distilled on-device student (red) achieves accuracy comparable to the optimized VFM teacher (blue) while requiring $336\times$ fewer parameters.}
    \label{fig:viz2d}
\end{figure}

\paragraph{Distillation with Optimized VFM.} The methods distilled from our optimized VFM yield substantial improvements over the baselines without VFM. \themethod{} achieves the best E50U50 performance, which reflects the typical user experience during normal conditions. Community KD has the best performance on E75U75, and training with pseudo labels only achieves the best E90U90. We note that tail samples in crowd-sourced datasets are more susceptible to label noise, arising from ambiguous gaze targets or inconsistent collection conditions, which may disproportionately affect tail metrics. In this regime, directly regressing to predictions from a strong optimized teacher is already a robust objective, which limits the marginal gains from additional distillation losses. With an optimized strong teacher, student performance may be driven more by teacher quality than the specific distillation objective. In scenarios where a strong teacher is not available, our proposed method provides more robust performance gains than these baselines, as shown in Supplementary \cref{sec:supp.metrics.dist_student} with a ConvNeXt-S teacher.

Compared to the baseline model, on-device synthetic supervision, \themethod{} reduces E50U50 from 3.48$^\circ$ to 1.44$^\circ$ (58.6\% improvement). Performance improvements also extend to the tail: E90U90 decreases from 14.84$^\circ$ to 8.45$^\circ$ (43.1\% reduction). We show population coverage charts at error percentiles E50, E75, and E90 in \cref{fig:coverage}, which visualize the complete distribution between users. Our optimized VFM and \themethod{} student exhibit closely matched performance across all user percentiles at E50 and E75, with a modest degradation at E90, demonstrating strong knowledge transfer during distillation. Both methods show a strong performance approaching a fully supervised on-device model trained with labeled data that would not be available during product development (last row in \cref{tab:results.distill}). Qualitative comparisons are presented in \cref{fig:viz2d}.

\subsection{Ablation Studies} \label{sec:results.ablate}

\begin{wrapfigure}[22]{r}{0.5\textwidth}
    \centering
    \vspace{-1em}
    \includegraphics[width=\linewidth]{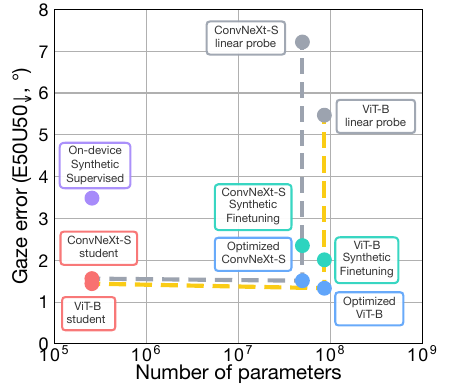}
    \captionof{figure}{\textbf{Ablations on VFM architecture.} Gaze error (E50U50 $\downarrow$, $^\circ$) vs. parameter count for ConvNeXt-S and ViT-B backbones. Linear probing (gray) performs poorly despite high parameter counts. Synthetic fine-tuning and our optimization substantially reduce error. Distilled on-device students (left) approach optimized VFM accuracy with orders of magnitude fewer parameters.}
    \label{fig:arch_compare}
\end{wrapfigure}

\paragraph{Teacher loss function.}
Many conventional self-supervised methods employ centering and prototype-based soft clustering to prevent representation collapse~\cite{caron2021emerging, oquab2023dinov2}. However, we find that a simple MSE regression loss substantially outperforms the DINO objective in our setting (\cref{tab:results.loss.ablate}). We attribute this to two key factors. First, soft clustering encourages discrete, categorical representations that may limit the fine-grained resolution required for continuous regression tasks like gaze estimation. Second, our use of synthetic supervision provides an inherent regularization signal that prevents collapse, reducing the reliance on these explicit mechanisms employed in self-supervised settings.

\begin{table}[t!]
    \centering
    \caption{\textbf{Ablations: teacher loss function.}  95\% confidence intervals in parentheses.}
    \label{tab:results.loss.ablate}
    \setlength{\tabcolsep}{1ex}
    \begin{tabular}{lccc}
        \toprule
        & E50U50 $\downarrow$ & E75U75 $\downarrow$ & E90U90 $\downarrow$ \\
        \midrule
        Optimized VFM {\color[HTML]{2563eb}$\blacktriangle$}  & \textbf{1.33 (0.07)} & \textbf{3.07 (0.34)} & \textbf{8.32 (2.10)} \\
        Optimized VFM {\color[HTML]{2563eb}$\blacktriangle$} using DINO loss & 1.50 (0.10) & 3.32 (0.38) & 8.86 (1.94) \\
        \bottomrule
    \end{tabular}
\end{table}

\paragraph{VFM model architecture.} One of our hypotheses was that ConvNeXt models would perform better than ViTs, since eye tracking features are concentrated near the pupil, and model performance may be constrained by ViT tokenization resolution. Interestingly, as we can observe in \cref{fig:arch_compare}, this is not the case. ViT-B exhibits stronger linear probe performance, and this advantage propagates through synthetic fine-tuning and self-distillation. However, the performance gap progressively narrows. In particular, both VFMs substantially outperform the on-device model with synthetic supervision baseline, which we attribute to the larger capacity of the models and the strong priors inherited from DINOv3 initialization. We speculate that peripheral regions of the eye visible within each patch may provide sufficient contextual cues, enabling ViT to match or exceed ConvNeXt's performance.

\subsection{Limitations and Future Work}
Although our method substantially improves over on-device baselines, a gap 
remains to the fully supervised upper bound, which uses labeled data that would not 
be available during product development. This gap is larger in the tail, 
suggesting that challenging cases, such as users with atypical eye appearance or 
extreme gaze angles, remain difficult without labeled supervision. 

Several promising directions may help narrow this gap, including bridging the domain discrepancy between real and synthetic data through more advanced generative modeling, designing stronger self-supervised objectives, or incorporating lightweight online adaptation at inference time.
In addition, all training and evaluation in this present work are conducted on a single device configuration. Extending the proposed framework to multiple devices with varying 
camera geometries, illumination, and sensor characteristics is an important 
direction for future work.

\section{Conclusion}
We presented \themethod{}, a framework for the rapid deployment of on-device eye tracking systems that leverages visual foundation models without requiring real ground-truth gaze supervision. Our analysis reveals that while DINOv3 representations cluster by subject identity, intra-subject distributions exhibit smooth gaze-correlated structure amenable to adaptation. We exploit this finding through a two-stage approach: first, optimizing DINOv3 via self-distillation combined with synthetic supervision and unlabeled real data, and then distilling the result into a compact on-device model. The final student achieves 58.6\% and 43.1\% reductions in E50U50 and E90U90, respectively, over a synthetic-supervised baseline, without any real ground-truth labels and with no increase in model size. By eliminating the need for labeled real data, \themethod{} enables rapid adaptation of the model to new hardware configurations, reducing the development cycle from months to days.

\bibliographystyle{include/eccv/splncs04}
\bibliography{main}

\clearpage
\appendix
\section{Additional Data Description}
\begin{figure}[h!]
    \centering
    \includegraphics[width=\linewidth]{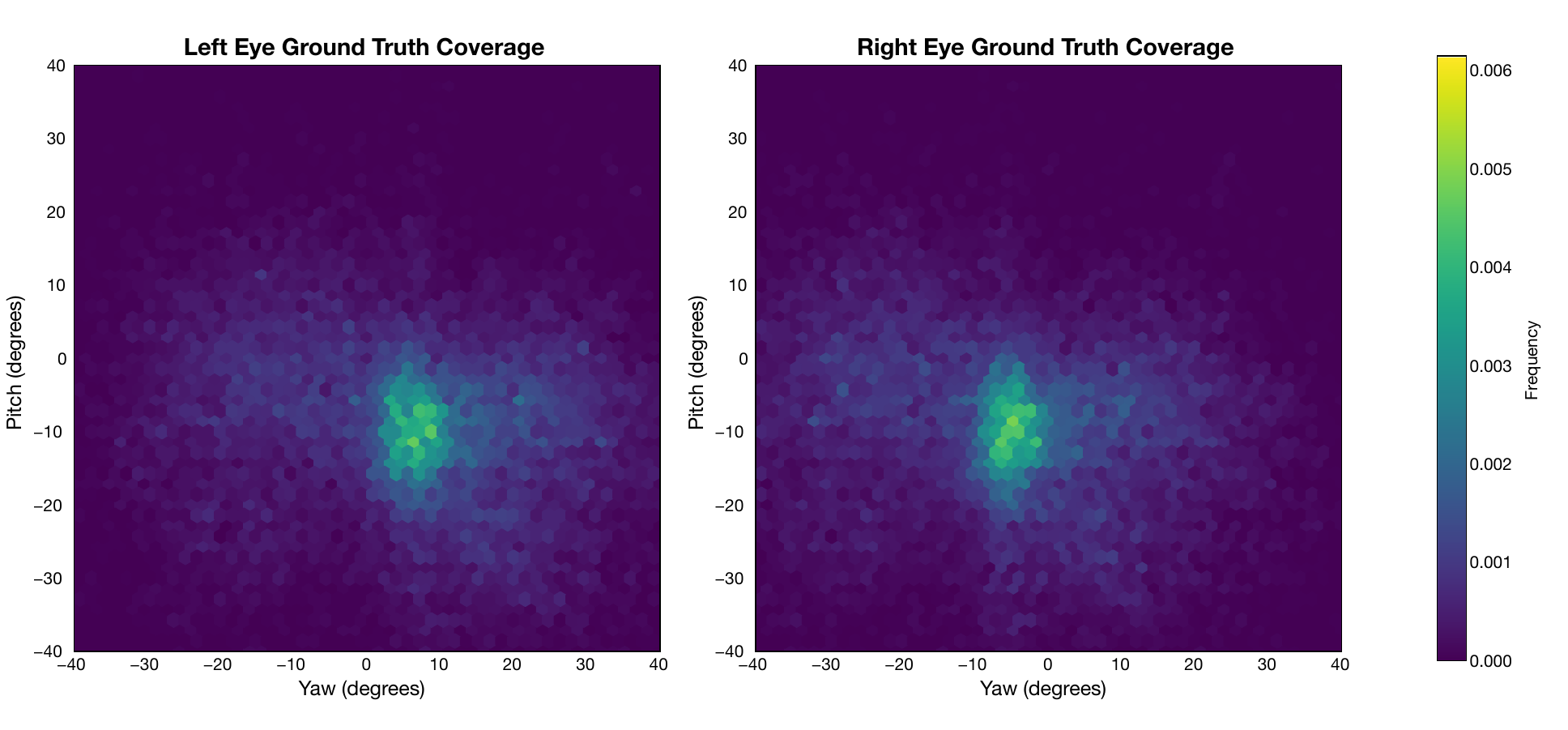}
    \caption{Project Aria benchmark ground truth distribution}
    \label{fig:aria.gt}
\end{figure}

\begin{figure}[h!]
    \centering
    \includegraphics[width=\linewidth]{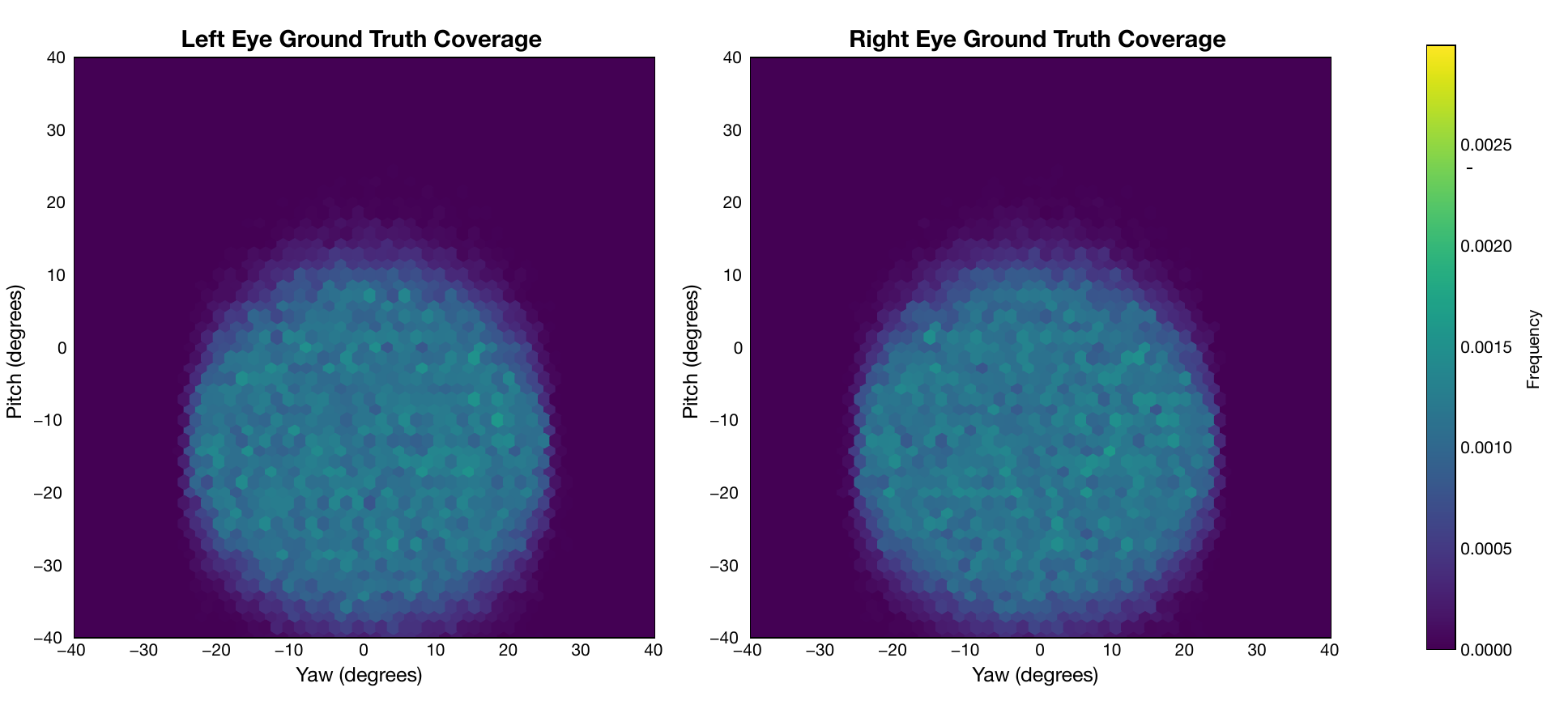}
    \caption{Project Aria synthetic data ground truth distribution}
    \label{fig:aria.gt_syn}
\end{figure}

In \cref{exp.data}, we described the Project Aria dataset. The distribution of the real data ground truth gaze is shown in \cref{fig:aria.gt}, and the distribution of the synthetic data ground truth is shown in \cref{fig:aria.gt_syn}. During evaluation, data points with yaw or pitch angles greater than $40^\circ$ or less than $-40^\circ$ are removed. The synthetic data is generated with uniform ground truth distribution over a wide variety of possible gaze angles, covering the most frequent ground truth in the crowd-sourced Project Aria set.

\clearpage
\section{Extended Results}

In this section, we present extended results from the manuscript, including full EU tables computed at user percentiles U50, U75, U90, and error percentiles E50, E75, E90.

\subsection{DINOv3 Embedding Visualization with Gaze Ground Truth}
\begin{figure}[h!]
    \centering
    \includegraphics[width=\textwidth]{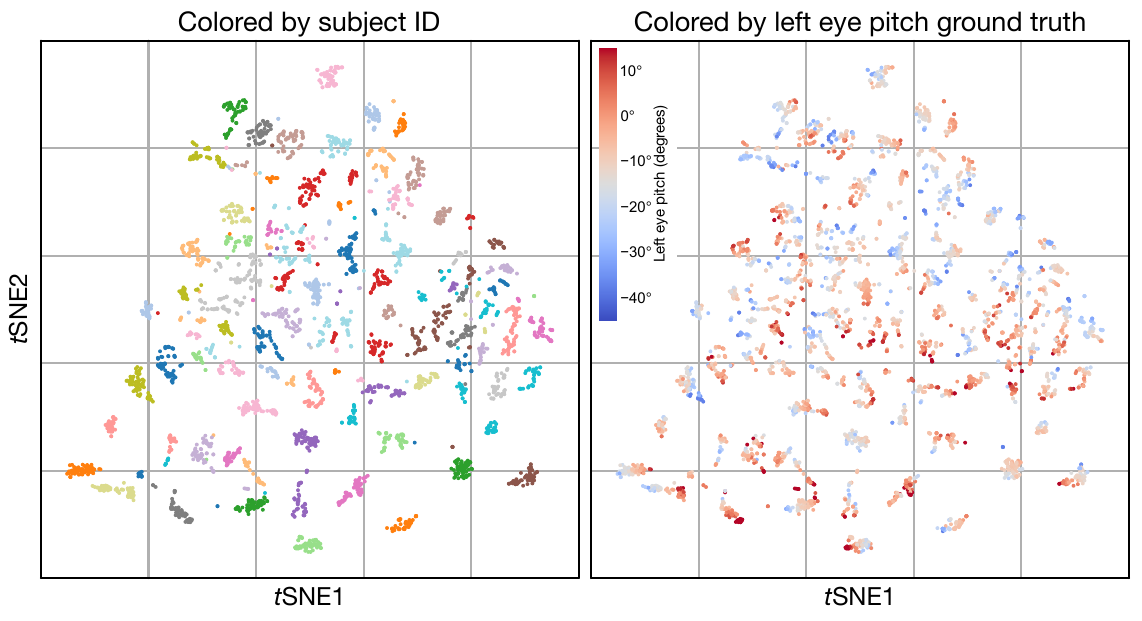}
    \caption{\textbf{Complete t-SNE visualization of DINOv3 embeddings colored by gaze ground truth.} DINOv3 embeddings for near-eye images form clear subject-level clusters, confirming that subject identity primarily determines the global representation structure. At the same time, the gradual progression of gaze values within each cluster reveals locally coherent organization with respect to gaze direction, indicating that the representation also preserves meaningful local gaze structure.}
    \label{fig:supp.tsne_gt_full}
\end{figure}

In \cref{fig:motivation.tsne} we presented DINOv3 embeddings on ET images, and discovered global subject clusters, with intra-subject distributions showing smooth gaze correlated structure. We present the complete visualization colored by gaze ground-truth in \cref{fig:supp.tsne_gt_full}.

\clearpage
\subsection{Extended VFM Linear probing Analysis}

\begin{figure}[h!]
    \centering
    \includegraphics[width=\textwidth]{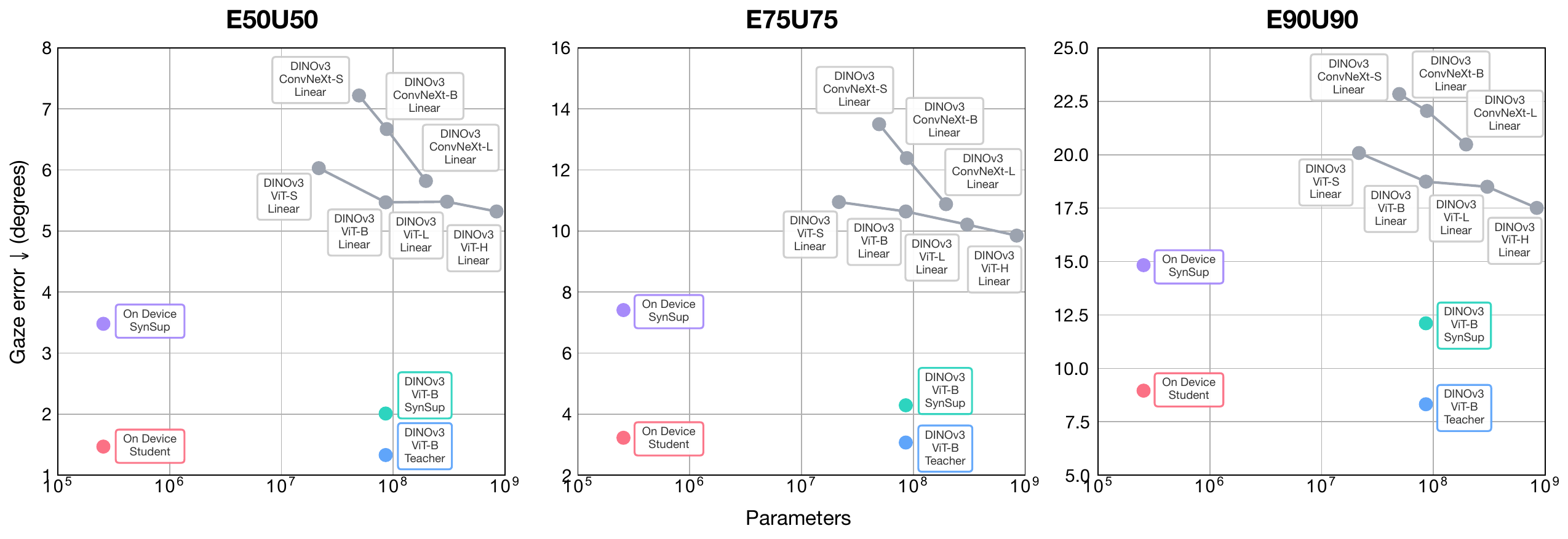}
    
    \caption{\textbf{Comparing Full EU table of DINOv3 family linear probe with our optimized VFM and on-device models.} DINOv3 family models with synthetic supervision (gray) significantly underperforms our on-device model trained with synthetic supervision (purple) in all three metrics reported. Our approach combines self-distillation with synthetic supervision to optimize a VFM using the ViT-B architecture (blue), and distill it into an on-device model with matching performance (red).
}
    \label{fig:supp.full_teaser}  
\end{figure}

In \cref{fig:motivation.overview} and \cref{tab:results.main}, we presented linear probing analysis on the DINOv3 family VFMs. We report the full EU table in \cref{tab:extended.linear}. ViT-H+, the largest evaluated model in the DINOv3 family with 840M parameters, has the best performance, but it still underperforms our on-device model with 256K parameters. An extended \cref{fig:motivation.overview} with E75U75 and E90U90 is shown in \cref{fig:supp.full_teaser}. The observations that motivate \themethod{} from \cref{fig:motivation.overview} generalize across E75U75 and E90U90 metrics.

\begin{table}[h!]
\centering\caption{\textbf{Full EU table for DINOv3 linear probe.} 95\% confidence intervals reported in parentheses.  On-device SynSup: on-device model trained with synthetic data supervision provided for reference. Inf. params: number of parameters at inference time.}\label{tab:extended.linear}%
\resizebox{\textwidth}{!}{%
\begin{tabular}{lc ccc ccc ccc}
\toprule
&&\multicolumn{3}{c}{U50 $\downarrow$} & \multicolumn{3}{c}{U75 $\downarrow$} & \multicolumn{3}{c}{U90 $\downarrow$} \\
\cmidrule(lr){3-5} \cmidrule(lr){6-8} \cmidrule(lr){9-11}
& Inf. params & E50 $\downarrow$ & E75 $\downarrow$ & E90 $\downarrow$ & E50 $\downarrow$ & E75 $\downarrow$ & E90 $\downarrow$ & E50 $\downarrow$ & E75 $\downarrow$ & E90 $\downarrow$\\
\midrule
ConvNext-S                                             & 50  M & 7.22 (0.33) & 10.84 (0.48) & 15.31 (0.72) & 9.04 (0.42) & 13.50 (0.59) & 18.81 (0.83) & 10.96 (0.69) & 16.79 (1.06) & 22.84 (1.44)\\
ConvNext-B                                             & 89  M & 6.67 (0.26) & 9.97 (0.41)  & 13.56 (0.58) & 8.18 (0.37) & 12.39 (0.67) & 17.48 (0.94) & 10.12 (0.72) & 15.25 (1.00) & 22.05 (1.67)\\
ConvNext-L                                             & 198 M & 5.82 (0.28) & 8.54 (0.39)  & 11.62 (0.49) & 7.56 (0.39) & 10.88 (0.58) & 15.02 (0.92) & 9.05 (0.64)  & 13.99 (1.09) & 20.40 (2.19)\\\midrule
ViT-S                                                  & 21  M & 6.03 (0.28) & 8.76 (0.37)  & 11.86 (0.62) & 7.44 (0.35) & 10.95 (0.60) & 15.75 (0.83) & 9.32 (0.61)  & 13.78 (1.05) & 20.08 (1.58)\\
ViT-B                                                  & 86  M & 5.47 (0.23) & 7.98 (0.37)  & 11.13 (0.56) & 6.95 (0.41) & 10.64 (0.61) & 14.57 (0.93) & 8.90 (0.64)  & 13.31 (1.01) & 18.74 (1.60)\\
ViT-L                                                  & 300 M & 5.48 (0.24) & 7.86 (0.34)  & 10.79 (0.50) & 6.80 (0.41) & 10.21 (0.63) & 14.17 (0.83) & 8.61 (0.61)  & 13.27 (1.18) & 18.50 (1.66)\\
ViT-H+                                                 & 840 M & 5.32 (0.24) & 7.82 (0.35)  & 10.68 (0.47) & 6.74 (0.32) & 9.85 (0.49)  & 13.41 (0.77) & 8.16 (0.54)  & 12.57 (1.07) & 17.51 (1.45)\\\midrule
On-device SynSup {\color[HTML]{7c3aed}$\blacklozenge$} & 256 K & 3.48 (0.19) & 5.30 (0.30)  & 7.87 (0.48)  & 4.82 (0.34) & 7.41 (0.53)  & 10.69 (0.83) & 6.56 (0.62)  & 10.48 (1.07) & 14.84 (2.03)\\\bottomrule
\end{tabular}
}
\end{table}

\clearpage
\subsection{Extended Metrics for Optimized VFM}

In \cref{tab:results.main}, we presented metrics for our optimized VFM. We report the full EU results in \cref{tab:extended.ovfm}. In addition, we also included the same experiments with a ConvNeXt-S backbone. Optimized VFM outperforms all baselines for both ViT-B and ConvNeXt-S backbones. Comparing architectures, ViT-B consistently outperforms ConvNeXt-S under matched training protocols, suggesting that the larger backbone and stronger VFM priors remain beneficial after adaptation to eye tracking. We also observe that DARE-GRAM~\cite{nejjar2023dare}, a state-of-the-art unsupervised domain adaptation baseline, underperforms our synthetic finetuning baseline in this setting. Moreover, its performance is stronger for smaller models, with the best results appearing for the on-device model (shown in \cref{tab:extended.dist}). These results suggest that DARE-GRAM is less effective for larger backbones in our synthetic-to-real adaptation setting. A possible explanation is that its adaptation objective may impose synthetic-domain structure that is not well aligned with the gaze distribution in real data, limiting downstream performance.

\begin{table}[h!]
    \centering\caption{\textbf{Gaze estimation performance comparing linear-evaluated foundation models and models optimized with synthetic and unlabeled real data.} 95\% confidence intervals reported in parentheses. Inf. params: number of parameters at inference, On-device SynSup: on-device model trained with synthetic data supervision provided for reference. }\label{tab:extended.ovfm}%
    \resizebox{\textwidth}{!}{%
\begin{tabular}{lcc ccc ccc ccc}
\toprule
&&&\multicolumn{3}{c}{U50 $\downarrow$} & \multicolumn{3}{c}{U75 $\downarrow$} & \multicolumn{3}{c}{U90 $\downarrow$} \\
\cmidrule(lr){4-6} \cmidrule(lr){7-9} \cmidrule(lr){10-12}
& Architecture & Inf. params & E50 $\downarrow$ & E75 $\downarrow$ & E90 $\downarrow$ & E50 $\downarrow$ & E75 $\downarrow$ & E90 $\downarrow$ & E50 $\downarrow$ & E75 $\downarrow$ & E90 $\downarrow$\\
\midrule
On-device SynSup {\color[HTML]{7c3aed}$\blacklozenge$}  & FB-Net     & 256 K & 3.48 (0.19) & 5.30 (0.30) & 7.87 (0.48) & 4.82 (0.34) & 7.41 (0.53) & 10.69 (0.83) & 6.56 (0.62) & 10.48 (1.07) & 14.84 (2.03)\\\midrule
DINOv3 \cite{simeoni2025dinov3} linear probe            & ViT-B      & 86 M & 5.47 (0.23) & 7.98 (0.37) & 11.13 (0.56) & 6.95 (0.41) & 10.64 (0.61) & 14.57 (0.93) & 8.90 (0.64) & 13.31 (1.01) & 18.74 (1.60)\\
DINOv3 \cite{simeoni2025dinov3} synthetic finetuning    & ViT-B      & 86 M & 2.01 (0.11) & 3.06 (0.18) & 4.51 (0.28) & 2.79 (0.32) & 4.29 (0.46) & 7.16 (0.84) & 4.92 (0.75) & 7.80 (1.07) & 12.11 (1.50)\\
DARE-GRAM \cite{nejjar2023dare}                         & ViT-B      & 86 M & 4.84 (0.24) & 7.40 (0.34) & 10.79 (0.52) & 6.32 (0.33) & 9.36 (0.54) & 13.84 (0.73) & 8.25 (0.65) & 12.21 (1.11) & 17.89 (1.65)\\
Optimized VFM {\color[HTML]{2563eb}$\blacktriangle$}    & ViT-B      & 86 M & \textbf{1.33 (0.07)} & \textbf{2.03 (0.12)} & \textbf{3.10 (0.22)} & \textbf{1.95 (0.18)} & \textbf{3.07 (0.34)} & \textbf{4.70 (0.50)} & \textbf{3.39 (0.64)} & \textbf{5.28 (0.90)} & \textbf{8.32 (2.10)}\\\midrule
DINOv3 linear probe                                     & ConvNeXt-S & 49 M & 7.22 (0.33) & 10.84 (0.48) & 15.31 (0.72) & 9.04 (0.42) & 13.50 (0.59) & 18.81 (0.83) & 10.96 (0.69) & 16.79 (1.06) & 22.84 (1.44)\\
DINOv3 synthetic finetuning                             & ConvNeXt-S & 49 M & 2.35 (0.14) & 3.57 (0.19) & 5.17 (0.35) & 3.37 (0.29) & 5.10 (0.48) & 7.73 (0.67) & 4.93 (0.64) & 7.99 (1.02) & 11.85 (1.60)\\
DARE-GRAM \cite{nejjar2023dare}                         & ConvNeXt-S & 49 M & 3.13 (0.15) & 4.83 (0.25) & 7.27 (0.42) & 4.30 (0.29) & 6.61 (0.50) & 10.38 (0.93) & 6.01 (0.67) & 9.45 (1.10) & 14.86 (1.67)\\
Optimized VFM                                           & ConvNeXt-S & 49 M & \textbf{1.51 (0.08)} & \textbf{2.33 (0.13)} & \textbf{3.65 (0.27)} & \textbf{2.15 (0.22)} & \textbf{3.39 (0.27)} & \textbf{5.20 (0.48)} & \textbf{3.38 (0.64)} & \textbf{5.49 (0.92)} & \textbf{8.62 (1.81)}\\\bottomrule
\end{tabular}
}
\end{table}

\clearpage
\subsection{Extended Metrics for Distilling On-Device Student}\label{sec:supp.metrics.dist_student}

In \cref{sec:results.distill} and \cref{tab:results.distill},  we presented metrics for our distilled student. We report the full EU table in \cref{tab:extended.dist}, along with additional results using an optimized DINOv3 ConvNeXt-S as distillation teacher. Overall, distillation from the optimized VFM substantially improves over training the on-device model without a VFM teacher, with all teacher-based methods clearly outperforming methods without distillation from a teacher, confirming the value of transferring knowledge from the optimized VFM to the compact student. Among distillation-based methods, \themethod{} achieves the best performance on E50U50, which is most reflective of average user experience under typical conditions, while remaining competitive across all other metrics. In contrast, for the tail setting E90U90, the simplest pseudo-label only distillation achieves the strongest performance. One possible explanation is that tail samples in crowd-sourced eye-tracking data are more susceptible to label noise, and directly regressing to teacher predictions may be more robust than imposing additional inductive biases for a better representation during distillation. We also observe that \themethod{} is relatively more competitive with the weaker ConvNeXt-S teacher, suggesting that when the teacher is stronger, overall student performance may depend less on the distillation objective.

\begin{table}[h!]
    \centering\caption{\textbf{Full EU table for on-device model performance comparison.}  Fully supervised upperbound is provided as reference, where it is trained using data that would not be available at the time of product development. 95\% confidence intervals reported in parentheses, Inf. params: number of parameters at inference time, teacher arch: architecture of teacher model (if applicable).}\label{tab:extended.dist}%
    \resizebox{\textwidth}{!}{%
\begin{tabular}{lcc ccc ccc ccc}
\toprule
&&&\multicolumn{3}{c}{U50 $\downarrow$} & \multicolumn{3}{c}{U75 $\downarrow$} & \multicolumn{3}{c}{U90 $\downarrow$} \\
\cmidrule(lr){4-6} \cmidrule(lr){7-9} \cmidrule(lr){10-12}
& Inf. params & Teacher arch. & E50 $\downarrow$ & E75 $\downarrow$ & E90 $\downarrow$ & E50 $\downarrow$ & E75 $\downarrow$ & E90 $\downarrow$ & E50 $\downarrow$ & E75 $\downarrow$ & E90 $\downarrow$\\
\midrule
\multicolumn{5}{l}{\emph{Methods without VFM as teacher}}\\
On-device synthetic supervision {\color[HTML]{7c3aed}$\blacklozenge$} & 256 K & - & 3.48 (0.19) & 5.30 (0.30) & 7.87 (0.48) & 4.82 (0.34) & 7.41 (0.53) & 10.69 (0.83) & 6.56 (0.62) & 10.48 (1.07) & 14.84 (2.03)\\
DARE-GRAM \cite{nejjar2023dare}                                       & 256 K & - &2.96 (0.17) & 4.45 (0.26) & 6.53 (0.35) & 4.08 (0.37) & 6.20 (0.47) & 9.28 (0.67) & 6.15 (0.69) & 9.20 (1.10) & 13.41 (1.73)\\
Self-distillation                                                     & 256 K & - &2.20 (0.11) & 3.22 (0.19) & 4.63 (0.30) & 3.03 (0.31) & 4.64 (0.46) & 7.16 (0.79) & 5.14 (0.68) & 7.82 (1.01) & 10.95 (2.10)\\\midrule
\multicolumn{5}{l}{\emph{Distillation with Optimized VFM (ViT-B) as teacher}}\\
Optimized VFM (ViT-B) {\color[HTML]{2563eb}$\blacktriangle$} & 86 M             & -     &1.33 (0.07) & 2.03 (0.12) & 3.10 (0.22) & 1.95 (0.18) & 3.07 (0.34) & 4.70 (0.50) & 3.39 (0.64) & 5.28 (0.90) & 8.32 (2.10)\\
Pseudo labels                                                & 256 K            & ViT-B & 1.59 (0.09) & 2.47 (0.14) & 3.86 (0.25) & 2.34 (0.20) & 3.49 (0.30) & 5.70 (0.48) &  \textbf{3.54 (0.68)} & 5.47 (0.75) &  \textbf{8.19 (1.56)}\\
SP \cite{tung2019similarity}                                 & 256 K            & ViT-B & 1.59 (0.08) & 2.48 (0.14) & 3.89 (0.23) & 2.32 (0.22) & 3.50 (0.29) & 5.68 (0.48) & 3.64 (0.66) & 5.52 (0.74) &  \underline{8.21 (1.71)}\\
VIC-KD \cite{guimaraes2024vic}                               & 256 K            & ViT-B & \underline{1.46 (0.08)} &  \textbf{2.20 (0.15)} &  \textbf{3.47 (0.29)} & \underline{2.11 (0.21)} & 3.32 (0.35) & 5.42 (0.51) &  \textbf{3.54 (0.49)} &  \textbf{5.36 (0.90)} & 8.40 (2.01)\\
Community KD \cite{lee2023co}                                & 256 K            & ViT-B & 1.48 (0.09) & \underline{2.21 (0.15)} &  \textbf{3.47 (0.27)} & 2.12 (0.21) &  \textbf{3.24 (0.31)} &  \textbf{5.14 (0.53)} &  \textbf{3.54 (0.55)} &  \textbf{5.34 (0.93)} & 8.47 (1.88)\\
\themethod{} (ours) {\color[HTML]{dc2626}$\bigstar$}         & 256 K            & ViT-B &  \textbf{1.44 (0.09)} & 2.27 (0.15) & 3.48 (0.24) &  \textbf{2.06 (0.20)} & \underline{3.29 (0.32)} &  \underline{5.24 (0.56)} & 3.63 (0.62) & 5.44 (0.96) & 8.45 (2.04)\\\midrule
\multicolumn{5}{l}{\emph{Distillation with Optimized VFM (ConvNeXt-S) as teacher}}\\
Optimized VFM (ConvNeXt-S)                                   & 49 M             & -          & 1.51 (0.08) & 2.33 (0.13) & 3.65 (0.27) & 2.15 (0.22) & 3.39 (0.27) & 5.20 (0.48) & 3.38 (0.64) & 5.49 (0.92) & 8.62 (1.81)\\
Pseudo labels                                                & 256 K            & ConvNeXt-S & \underline{1.59 (0.09)} & 2.48 (0.16) & 3.81 (0.26) & \underline{2.32 (0.21)} & \underline{3.38 (0.28)} & 5.56 (0.47) & \textbf{3.58 (0.67)} & \underline{5.46 (0.64)} & \textbf{8.16 (1.62)}\\
SP \cite{tung2019similarity}                                 & 256 K            & ConvNeXt-S & 1.63 (0.09) & 2.51 (0.15) & 3.89 (0.25) & \underline{2.32 (0.21)} & 3.51 (0.27) & 5.77 (0.48) & \underline{3.61 (0.65)} & 5.48 (0.76) & 8.26 (1.67)\\
VIC-KD \cite{guimaraes2024vic}                               & 256 K            & ConvNeXt-S & 1.61 (0.09) & 2.46 (0.15) & 3.83 (0.27) & 2.38 (0.20) & 3.48 (0.25) & \underline{5.54 (0.45)} & 3.67 (0.73) & \underline{5.46 (0.84)} & \underline{8.17 (1.58)}\\
Community KD \cite{lee2023co}                                & 256 K            & ConvNeXt-S & 1.61 (0.08) & \underline{2.41 (0.15)} & \underline{3.77 (0.25)} & 2.33 (0.22) & 3.44 (0.31) & 5.65 (0.47) & 3.65 (0.68) & \textbf{5.38 (0.80)} & 8.26 (1.66)\\
\themethod{}                                                 & 256 K            & ConvNeXt-S & \textbf{1.56 (0.09)} & \textbf{2.36 (0.15)} & \textbf{3.67 (0.24)} & \textbf{2.29 (0.24)} & \textbf{3.34 (0.32)} & \textbf{5.44 (0.50)} & 3.74 (0.60) & 5.58 (0.84) & 8.75 (1.62)\\\midrule
\graycell{Fully supervised (upperbound)}                     & \graycell{256 K} & \graycell{-} & \graycell{0.82 (0.05)} & \graycell{1.16 (0.06)} & \graycell{1.66 (0.14)} & \graycell{1.26 (0.18)} & \graycell{1.94 (0.26)} & \graycell{2.92 (0.47)} & \graycell{2.49 (0.73)} & \graycell{3.66 (1.13)} & \graycell{5.91 (1.61)}\\
\bottomrule
\end{tabular}
}
\end{table}

\clearpage
\subsection{Additional Prediction Visualizations}

\begin{figure}[h!]
    \centering
    \includegraphics[width=\textwidth]{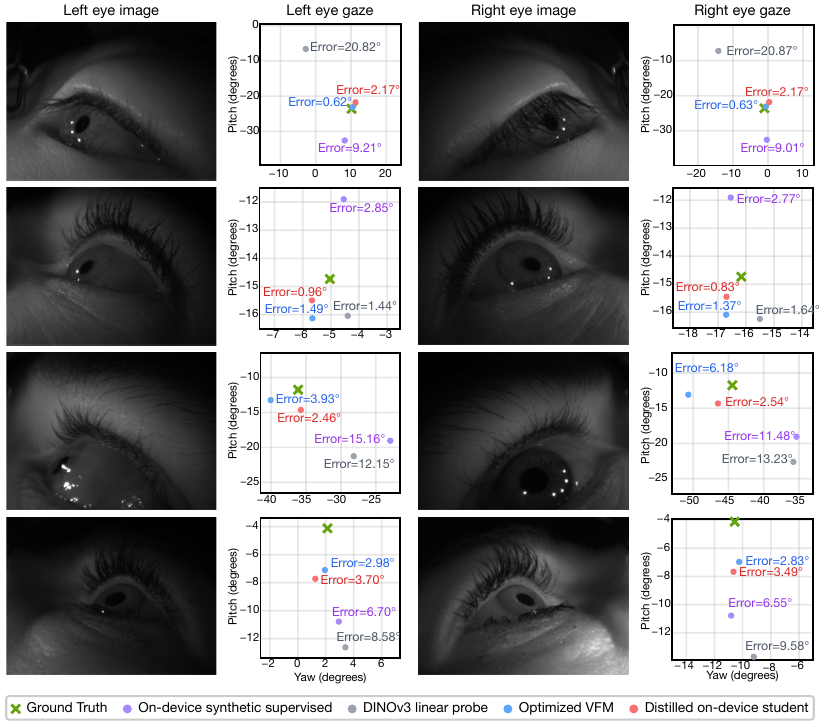}
    
    \caption{\textbf{Additional qualitative comparisons of gaze predictions.} DINOv3 linear probe (gray) and on-device synthetic supervised model (purple) yield relatively large gaze errors, while our optimized VFM (blue) significantly reduces the errors. Our distilled on-device student (red) shows similar performance to our optimized VFM, while using significantly fewer parameters.
    }
    \label{fig:supp.viz2d}
    
\end{figure}

We presented qualitative comparisons of gaze predictions using pitch-yaw plots in \cref{fig:viz2d}. We present additional visualizations in \cref{fig:supp.viz2d}. Both the DINOv3 linear probe and the on-device model trained with synthetic supervision yield relatively large gaze errors. In contrast, our optimized VFM substantially reduces the error, and the distilled on-device student achieves comparable performance to the optimized VFM while using significantly fewer parameters.

\clearpage

\subsection{Loss Function Ablations for Optimizing VFM}

In \cref{sec:results.ablate} and \cref{tab:results.loss.ablate}, we presented ablations on optimizing VFM features. We report the full EU table in \cref{tab:extended.loss.ablate}.

\begin{table}[h!]
\centering\caption{\textbf{Full EU table for loss function ablations.} Our self-distillation algorithm combined with synthetic supervision performs well with a simple MSE objective. 95\% confidence intervals reported in parentheses. Inf. params: number of parameters at inference time.}\label{tab:extended.loss.ablate}%
\resizebox{\textwidth}{!}{%
\begin{tabular}{lc ccc ccc ccc}
\toprule
&&\multicolumn{3}{c}{U50 $\downarrow$} & \multicolumn{3}{c}{U75 $\downarrow$} & \multicolumn{3}{c}{U90 $\downarrow$} \\
\cmidrule(lr){3-5} \cmidrule(lr){6-8} \cmidrule(lr){9-11}
& Inf. params & E50 $\downarrow$ & E75 $\downarrow$ & E90 $\downarrow$ & E50 $\downarrow$ & E75 $\downarrow$ & E90 $\downarrow$ & E50 $\downarrow$ & E75 $\downarrow$ & E90 $\downarrow$\\
\midrule
Optimized VFM {\color[HTML]{2563eb}$\blacktriangle$}    & 86 M & \textbf{1.33 (0.07)} & \textbf{2.03 (0.12)} & \textbf{3.10 (0.22)} & \textbf{1.95 (0.18)} & \textbf{3.07 (0.34)} & \textbf{4.70 (0.50)} & \textbf{3.39 (0.64)} & \textbf{5.28 (0.90)} & \textbf{8.32 (2.10)}\\
Optimized VFM {\color[HTML]{2563eb}$\blacktriangle$} using DINO loss  & 86 M & 1.50 (0.10) & 2.25 (0.13) & 3.39 (0.23) & 2.15 (0.23) & 3.32 (0.38) & 5.26 (0.66) & 3.82 (0.78) & 5.83 (1.14) & 8.86 (1.94)\\\bottomrule
\end{tabular}
}
\end{table}

\subsection{Loss Function Weight Ablations for Optimizing VFM}
As described in \cref{sec:method.alignment}, our loss function for optimizing the VFM consists of synthetic supervision, self-distillation, and pseudo-labels, with a cosine scheduler that emphasizes synthetic supervision at the beginning of training and gradually transitions toward self-distillation and pseudo-labels. In this section, we show additional ablation studies on the loss functions and weighting. The results in \cref{tab:extended.ovfm.losswt} confirm that the optimized VFM is based on the combination of all of our loss components and the scheduled weighting between them. Removing either synthetic supervision or the scheduler leads to worse performance across metrics, with larger error than linear probing, highlighting the importance of first adapting the VFM to the eye-tracking task through synthetic supervision, as naively applying self-distillation and pseudo-label losses can degrade the VFM representation. In contrast, removing self-distillation and pseudo-label losses yields results comparable to synthetic fine-tuning, suggesting that its benefit mainly appears when combined with synthetic supervision under the proposed schedule. Overall, these findings support our design choice of first adapting the VFM representation to the eye-tracking task through synthetic supervision, and then further refining it with self-distillation and pseudo-labels under a gradual transition in loss weighting.

\begin{table}[h!]
    \caption{\textbf{Loss function weight ablations for optimized VFM.} We ablate synthetic supervision, self-distillation, and the cosine loss-weight scheduler used to transition between them. 95\% confidence intervals reported in parentheses. SynSup: training with synthetic supervision, SD+PL: training with self-distillation and pseudo-label losses, LS: loss scheduler.}\label{tab:extended.ovfm.losswt}%
    \resizebox{\textwidth}{!}{%
\begin{tabular}{l ccc ccc ccc ccc}
\toprule
&&&&\multicolumn{3}{c}{U50 $\downarrow$} & \multicolumn{3}{c}{U75 $\downarrow$} & \multicolumn{3}{c}{U90 $\downarrow$} \\
 \cmidrule(lr){5-7} \cmidrule(lr){8-10} \cmidrule(lr){11-13}
& SynSup & SD+PL & LS & E50 $\downarrow$ & E75 $\downarrow$ & E90 $\downarrow$ & E50 $\downarrow$ & E75 $\downarrow$ & E90 $\downarrow$ & E50 $\downarrow$ & E75 $\downarrow$ & E90 $\downarrow$\\
\midrule
Optimized VFM {\color[HTML]{2563eb}$\blacktriangle$}      & \checkmark     & \checkmark & \checkmark    & \textbf{1.33 (0.07)} & \textbf{2.03 (0.12)} & \textbf{3.10 (0.22)} & \textbf{1.95 (0.18)} & \textbf{3.07 (0.34)} & \textbf{4.70 (0.50)} & \textbf{3.39 (0.64)} & \textbf{5.28 (0.90)} & \textbf{8.32 (2.10)}\\
Self distillation and pseudo-labels only                  &                & \checkmark &               & 9.53 (0.38) & 13.93 (0.50)& 19.14 (0.74) & 11.70 (0.52) & 16.88 (0.66) & 22.63 (1.01) & 13.23 (0.58) & 19.31 (1.09) & 27.90 (1.61)\\
Synthetic supervision only                                & \checkmark     &            &               & 2.22 (0.13) & 3.45 (0.21) & 5.10 (0.35)  & 3.11 (0.26)  & 4.69 (0.44)  & 7.43 (0.62)  & 4.54 (0.53)  & 7.08 (0.92)  & 10.34 (1.69)\\
Remove loss weight scheduler                              & \checkmark     & \checkmark &               & 8.73 (0.41) & 12.75 (0.45)& 17.10 (0.53) & 10.67 (0.40) & 15.08 (0.55) & 20.42 (0.87) & 12.31 (0.52) & 17.70 (1.01) & 24.54 (1.11)\\\bottomrule
\end{tabular}
}
\end{table}

\end{document}